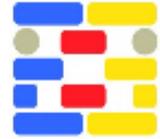

# TITLE BLOCK DETECTION AND INFORMATION EXTRACTION FOR ENHANCED BUILDING DRAWINGS SEARCH

Alessio Lombardi[1], Li Duan[2], Ahmed Elnagar[1], Ahmed Zaalouk[2], Khalid Ismail[2], Edlira Vakaj[2]
[1]Buro Happold Ltd., London, United Kingdom
[2]Birmingham City University, Birmingham, United Kingdom

## Abstract

The architecture, engineering, and construction (AEC) industry still heavily relies on information stored in drawings for building construction, maintenance, compliance and error checks. However, information extraction (IE) from building drawings is often time-consuming and costly, especially when dealing with historical buildings. Drawing search can be simplified by leveraging the information stored in the title block portion of the drawing, which can be seen as drawing metadata. However, title block IE can be complex especially when dealing with historical drawings which do not follow existing standards for uniformity. This work performs a comparison of existing methods for this kind of IE task, and then proposes a novel title block detection and IE pipeline which outperforms existing methods, in particular when dealing with complex, noisy historical drawings. The pipeline is obtained by combining a lightweight Convolutional Neural Network and GPT-4o, the proposed inference pipeline detects building engineering title blocks with high accuracy, and then extract structured drawing metadata from the title blocks, which can be used for drawing search, filtering and grouping. The work demonstrates high accuracy and efficiency in IE for both vector (CAD) and hand-drawn (historical) drawings. A user interface (UI) that leverages the extracted metadata for drawing search is established and deployed on real projects, which demonstrates significant time savings. Additionally, an extensible domain-expert-annotated dataset for title block detection is developed, via an efficient AEC-friendly annotation workflow that lays the foundation for future work.

## Introduction

Building drawings are a key element in the AEC industry. Historical drawings, in particular, are important for maintenance and refurbishment projects, although their quality is often poor and they are rarely aligned to modern conventions. Managing and retrieving information from technical drawings is time-consuming. Additionally, while several modern solutions have been proposed to address compliance checking on buildings and infrastructure, in many cases these processes still rely on exchange and manual commenting of drawings (Soliman-Junior et al. (2021)). Therefore, information extraction (IE) applied to building drawings is still very important in the AEC industry: a fast and effective method to search through a large number of drawings is crucial in the daily workflow of engineers and architects.

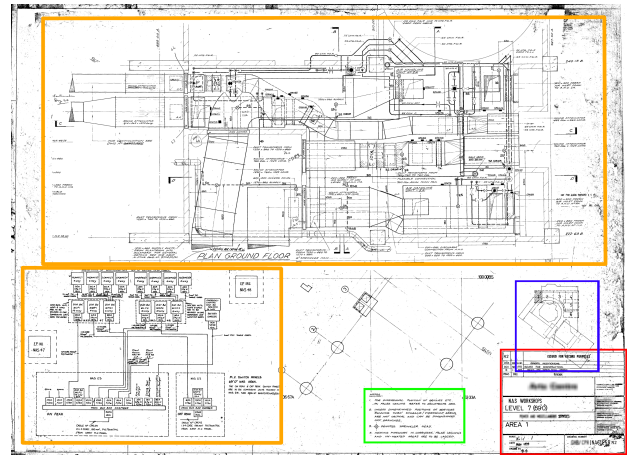

*Figure 1: An example of an engineering drawing demonstrating the different elements: title block (red), main drawing content (orange), legend (blue) and notes (green).*

However, drawing search is complex. Drawings aren't archived with conventions that allow simple retrieval. This varies depending on the project, and it's particularly true for building projects involving historical, pre-existing structures, such as refurbishment or extension projects. In these cases, the drawings are often only available as scans (of either computer-produced or hand-drawn drawings) and can be of poor quality. Searching specific drawings in such contexts can be a very tedious, manual task that can take several man-hours of work and is error-prone. On certain projects, in particular large refurbishment ones, a significant amount of the project budget (up to 40%) may be dedicated to this task alone. An example is the Barbican refurbishment project in London, where upwards of 42 thousand hand-drawn historical drawings produced in the 1970s need to be scoured for information. In general, this situation is common when dealing with buildings built before the 1990s, which constitute the vast majority of our built environment.

Technical drawings can be decomposed into several areas: *title block*, *main content*, and occasionally *legend* and *note* (Figure 1). The title block area contains drawing *metadata*, information about the drawing such as author, drawing number, scale, and/or information that describes the main content of the drawing itself. Thus, searching drawings based on the information in their title blocks is ideal, and is in fact what engineers typically do when needing to retrieve specific drawings. To extract information from title blocks, researchers such as Li et al. (2024); Kashevnik et al. (2023); Kang et al. (2019) proposed to apply traditional image processing technologies such as template match, sliding window and fuzzy matching to identify the area, then perform traditional OCR technique to extract relevant textual information.

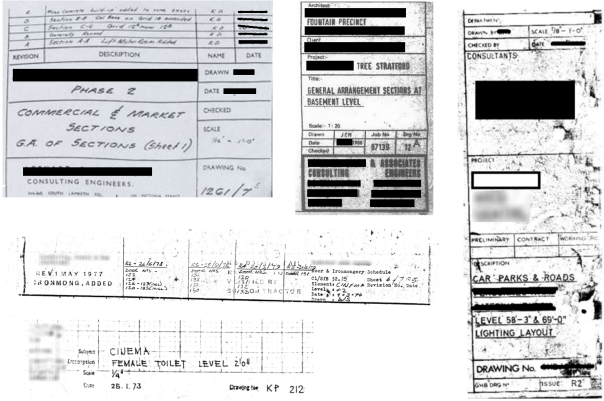

*Figure 2: Five examples of title blocks with different features. Sensitive information has been obscured.*

Classical, non-data-driven approaches to perform IE on title blocks use techniques that do not require large training data. However, these are not robust to variations in sizes, rotations, and in the case of historical or scanned drawings. Applying these approaches potentially results in errors when detecting the title block and extracting text from it. For these cases, the errors may be reduced by post-processing the extracted information when it is possible to rely on the assumption that the title blocks respect some kind of standard. However, while standards for title block exists, such as ISO 7200 and ASME Y14.1 (Li et al. (2024)), in practice there is a high variability in their adherence. Engineers frequently express concerns about inconsistent application of these standards, particularly in smaller to medium-sized firms or less regulated environments. Their strict application can vary depending on the project's scope, the organization's internal practices, and the specific requirements of clients. This is even more true when it comes to historical drawings, which are common in refurbishment projects. For example, the title blocks of the drawings of Figure 2 concern the same project but have significantly different styles.

Alternatively, other scientists (Shen et al. (2020); Smock et al. (2022); Ly and Takasu (2023)) explored data-driven approaches such as YOLO (You Only Look Once), Generative Adversarial Networks (GANs) and Transformers. Compared to traditional image processing technologies, these approaches are robust to symbol variations. However, extremely large datasets are needed to train and fine-tune these models, which are typically very difficult to obtain, especially due to sensitive information and data protection. Additionally, these approaches are expensive and difficult to implement due to their high computational cost.

The proposed approach overcomes the limitations of existing approaches by introducing a combined inference pipeline for title block detection and IE from building drawings. The title block detection is defined as a simple object detection task based on an ad-hoc dataset of labelled data developed for this work. The dataset contains building engineering drawings annotated by domain experts, and contains technical drawings with significant variations in title blocks and content. The dataset is used to fine-tune a FasterRCNN model (Ren et al. (2016)) to detect title blocks. Then, GPT-4o is used to extract structured information from the detected title blocks. This combined approach is demonstrably more efficient and effective than existing techniques. Additionally, the produced dataset serves as a foundational base that can be reused and expanded for future work related to computer vision (CV) tasks in engineering workflows.

The main contributions of this work are:

- test and compare existing approaches for IE from building drawings for drawing search;
- construct an extensible dataset of annotated building drawings using an efficient annotation workflow;
- propose an inference pipeline with FasterRCNN and GPT-4o for title block IE;
- deploy the pipeline and demonstrate drawing search through a simple ad-hoc UI.

The sections of the paper are as follows. *Related Work* discusses existing works and methodologies for IE from drawings. *Methodology, implementation and results* justifies the chosen approach via an initial comparative study on existing methods, explains the dataset and the proposed pipeline implementation, and discusses the results. *Conclusions, limitations and future work* illustrates the conclusions from this research and future research directions.

## Related Work

*Non-data-driven* and *data-driven* can be used to perform IE from table-like structures.

*Non-data-driven approaches* rely on classical machine learning approaches and/or heuristics to identify the structures of tables and extract information from them. In general, non-data-driven approaches are not very robust to changes and variations in table styles and symbol sizes.

Kang et al. (2019) explored template match and sliding window approaches to detect symbols and extract lines and texts in design drawings. Firstly, symbols are extracted by template match. Then, lines and texts are extracted using the sliding window approach. The extracted lines and symbols are associated with the attributes of the closest text information and stored in a dataset. However, symbols with small sizes often need to be included in their search.

Kashevnik et al. (2023) claimed that title block detection can be achieved by detecting their vertical and horizontal lines and intersections between these lines. From horizon-tal and vertical lines, they found rectangles that represent title blocks. They extracted information by splitting information blocks into cells. However, their approach was not effective on scanned or photographed drawings.

Li et al. (2024) proposed to extract title block information with fuzzy matching, where they evaluated the geometrical composition of title blocks, the logical relationship between information cells and the diversification of title blocks. They proposed a weighted intersection point matrix for title blocks based on the evaluation and implemented fuzzy matching based on the intersection point matrix. However, in their experiments, there were disorders within the arrangement and distribution of information, resulting in a certain number of empty cells and making extraction error-prone.

Works such as Van Daele et al. (2021) use a hybrid data-driven and non-data-driven approach, further expanding to neuro-symbolic IE techniques which combine the understanding of both the title block and the main drawing content. Limitations of such works is that they first expect the drawing elements (title block, main content) to be easily identifiable by assuming that a technical drawing employs white space to distinguish its elements. For example, Van Daele et al. (2021) proposes a simple DBSCAN algorithm to segment technical drawing in their constituting elements as in Figure 1. While this work states that no errors were observed in the segmentation of the drawings, their dataset includes only modern vector-based (i.e. produced with Computer Aided Design, CAD) drawings related to a particular engineering domains (mostly mechanical, electronics) where this assumption is generally true, because they are often tidier, less noisy and more regular than technical building drawings, especially historic ones such as the ones in Figure 1 and 2.

*Data-driven approaches* rely on large labelled datasets and Neural Networks (NN) to infer the location, structure and content of tables. They tend to generalise well and be more accurate, but they also require large models with a large amount of training data. Additionally, while they may generalise well for a certain type of tables, they may not generalise to hand-written tables and/or tables with a high variability in their content or shape.

Elyan et al. (2020) utilized YOLO Redmon (2016) and Deep Generative Adversarial Neural Network Goodfellow et al. (2014) to localize symbols in drawings and handle class imbalance within drawing symbols.

Shen et al. (2020) proposed layout parsing, which is a deep learning-based automatic document analysis approach for extracting information from historical documents. The approach demonstrated effectiveness and efficiency in extracting information from historical Japanese documents from training models on a large dataset of historical Japanese documents with complex layouts.

TableTransformer (Smock et al. (2022)) is a state-of-the-art (SOTA) approach to extracting texts from tables, especially columned tables in academic reports and papers. The strategy is based on transformers by training a deep-learning model on a large dataset called PubTables-1M. PubTables-1M is a dataset for table extraction from scientific articles, containing nearly one million tables. The inconsistencies in ground truth labels, which may arise when dealing with a large volume of data, with a novel canonicalization technique. This enhances training performance and provides reliable model evaluation. Transformer-based models trained on PubTables-1M have shown excellent performance in detection, structure recognition, and functional analysis without special customization.

MTL-TabNet (Ly and Takasu (2023)) is another SOTA table-extracting approach. MTL stands for Multi-task Learning-based Model for Image-Based Table Recognition. It is based on a shared encoder-decoder structure which splits the table recognition problem into 3 parts: table structure recognition, cell detection, and cell-content recognition. MTL-TabNet also shows satisfactory results on table extraction from publications.

GPT-4 (Achiam et al. (2023)) and GPT-4o can also be used for text extraction from structured or unstructured visual data. Specific details about the architecture, training methodology, and dataset construction of GPT-4 are not publicly available (?). However, the latest iteration of GPT-4o operates using a unified neural network trained end-to-end across text, vision, and audio. This integration allows the model to process all inputs and outputs within the same framework, enhancing its ability to understand and generate multimodal content. The model employs autoregressive transformers to handle sequential data, and its architecture includes mechanisms for compressing representations and composing autoregressive priors with powerful decoders. These features enable GPT-4o to achieve high performance in text, reasoning, and coding tasks. Unlike OCR, GPT-4o is able to understand the layout of tables and their content.

Azure AI Document Intelligence (AAIDI) is an online platform which provides a solution for extracting data from tables within documents. It utilises prebuilt models to with specific applications, such as tax reports and personal identification documents. AAIDI also includes options custom models for extraction and classification. This allows the user to label and build a custom model within AAIDI with a specific schema (Azure (2025)).

In summary, while traditional approaches such as template matching are simple to implement, they lack robustness against the variability and noise present in historical or scanned drawings. Conversely, state-of-the-art data-driven methods like transformers and GANs offer better performance but require large, domain-specific datasets that are costly to create. This paper addresses the gap between these extremes by proposing a lightweight yet effective pipeline that combines Faster RCNN (Ren et al.

(2016)) for title block detection and GPT-4o for IE. This approach is capable of handling diverse drawing types and conditions, and is trained on a relatively small, custom-labelled dataset, demonstrating its feasibility and adaptability for practical use in the AEC industry.

## Methodology, implementation and results

The scope of this study is to extract information from title block of building engineering drawings that can be used to perform useful search text-based queries. The main challenges that this problem poses are:

**(A)** *identification of a title block* within a technical engineering drawing; and

**(B)** *extraction of the information from the title block in a structured format* that can be usable for searching the related drawing.

Both task (A) and (B) can be performed in a number of ways. Different existing methods may work well for one of the two points, but can be inefficient or inaccurate when dealing with both, especially when applied to historical building engineering drawings. In particular, title blocks cells have generally a "cell title" that describes what kind of information is present in the cell, typically in the form of a smaller textual element in the top-left corner of the cell. Ideally, the output of task (B) would have the information extracted in some form of *key-value pair* structure, where each *cell value* is paired with the title of the cell (the *key*). This because the extracted information can then be easily treated as *drawing metadata*, minimizing the post-processing required for enabling drawing search. For example, the ideal text extraction from a drawing title block would look similar to:

```
{
    "Project name" : "Flat Iron",
    "Content" : "L03 kitchen section",
    "Date" : "Apr 1970",
    "Drawn by" : "GM, BF"
}
```

Obtaining the information from the title block in this structured *key-value pair metadata format* would minimize the post-processing required to implement search methods based on keywords. For example, 'Date' being equal to '1970' or 'Project name' containing 'Iron'. The extracted key-value pairs can always be grouped by key given their semantic meaning or with ad-hoc rules. For example, "Drawing content", "Drawing description" and "dwg desc" can be recognised as representing the same type of information. The extracted key-value pairs can then be paired with the drawing file, enabling search in existing File Systems or in common industry file platforms like SharePoint, or by developing a simple ad-hoc User Interface (UI).

This study evaluated existing methods and, based on the results, defined a novel pipeline able to accurately perform both task (A) and (B).

**Experiments with existing methods**

Initially a comparative study of existing methods was performed. This comparative study covered different methodologies and was used to inform the next steps of this work. The study used a test sample of 10 building engineering drawings selected by domain experts to be representative of different AEC domains and of difficult IE retrieval cases. The test sample included both vector (CAD) and historical (hand-drawn, scanned) drawings. In all cases, the drawings are treated as raster images in high resolution (300 dpi). The selected test sample contained variations in: a) complex title block shapes, b) irregular title block cells, c) variations in font size, d) variations in printed vs hand-written text, e) vertical/horizontal orientation, and f) mixed content in the title block cells (text only or text + images). Non-data-driven and data-driven methods were considered.

Among the *non-data-driven methods*, DBSCAN, template match and adaptive thresholding methods were tested to verify their ability of detecting the title block (task A). The reasoning is that if the drawing is clean enough, the title block can be easily recognised and is divided in its in sub-components, and its content can be then extracted using OCR techniques. These methods can be applied to tables (Hashemi et al. (2016)), and title blocks can be seen as a special case of a table. The Open-CV library (opencv.org) was used to test DBSCAN, template matching and adaptive thresholding on the small test dataset. The DBSCAN approach as proposed by Van Daele et al. (2021) returned very poor results when applied to cases such as those in Figure 1 and 2. Similarly, also the other methods failed. In at least 8 samples out of 10 for every method, the accuracy was below 15%. The accuracy was determined by defining a IoU threshold of 70% between the ground-truth title block cells and the extracted title block cells. This is very likely due to the high irregularity of the title blocks and of their scan quality. Regarding task B, text extraction techniques applied to title block was also poor, worsened by the imprecision of classical OCR techniques with noisy data.

*Data-driven approaches* for tables detection and IE were also tested. MTL-TabNet (Ly and Takasu (2023)) and TableTransformer (Smock et al. (2022)) showed a similar inability in both task (A) and (B). Both are trained on PubTables-1M, which has a different image domain from the title blocks in our experiments, and the tables in the dataset do have some regularity which is generally absent in our test set. For example, title blocks have variable rows/columns structures, and in many cases it is not even possible to talk about rows or columns, as their shape is deeply irregular. The results from both show poor splitting of the title block into different areas, which is likely due to the differences between title blocks and typical tabular structures. It is possible that further fine-tuning using an ad-hoc dataset would have lead to improved results. How-

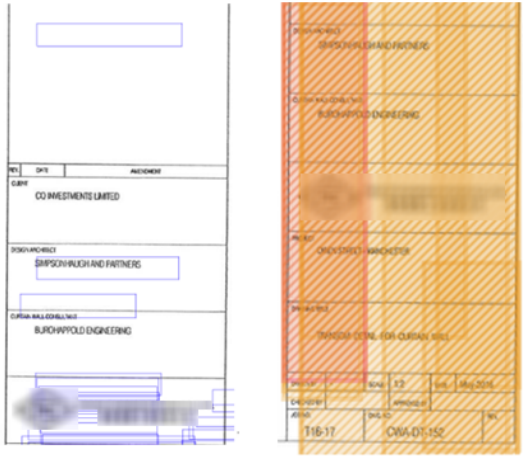

*Figure 3: Sample result from the title block splitting using existing deep learning models. Left: MTL-TabNet. Right: TableTransformer.*

```
"Drawng Title": "SECTION",
"Drawing Note 7": "JT3 5.5x25 THROUG SLOT",
"Drawing Note 8": "JT3 5.5x25 TO LOCK BOX",
"Drawing Note 10": "STEEL BY OTERS",
"Drawing Note 12": "3mm ALUMNIM BRCKET",
"Drawing No.": "A150",
"Checked by": "GR",
"Date" : "082014",
"Scale": "110",
"Rev": "01",
"Scale": "110"
```

*Figure 4: Example of title block text extraction using GPT-4o information from the entire drawing. The text misses important title block information, includes noise from outside the title block and has typos.*

```
"Drawing Title": "SECTION LEVEL 00",
"Drawing No.": "A1/50",
"Scale": "1:10",
"Date": "28.03.22",
"Drawn": "CJ",
"Checked": "CD",
"Rev": "P01",
"Status": "ISSUED FOR CONSTRUCTION",
"Drg. No.": "A1/50",
"Scale": "1:10",
"Drawn By": "CJ",
"Checked By": "CD"
```

*Figure 5: Example of title block text extraction on the same drawing of 4 using GPT-4o only on the title block portion of the drawing. The text is now complete and specific to the title block only and has no typos.*

ever, these methods focus on plain data extraction from table-like structures, extracting the text as-is. This does not help particularly with task (B), as the output would require a significant amount of post-processing to group the title block cells into key-value pairs. For task B, text extraction using Transformer-based Optical Character Recognition with Pre-trained Models (TrOCR, Li et al. (2023)) was also tested, but did not return satisfactory results either.

Subsequently, GPT-4o was tested. By performing prompt engineering, we tested the ability of GPT-4o to perform both title block detection (task A) and IE from the title block (task B). For both task A and B, the following system prompt was used: *You are an expert extracting text from an input image that is technical drawing*. For task A, the following prompt was used: *What text is in the titleblock of this drawing? Identify the title block with its cells, and return its contents in a JSON format, where the key is the cell title and the value is the content of the cell without the key.*

The tests demonstrated that GPT-4o was able to extract the text from the title block and output it into a structured JSON-like output. However, GPT4-o demonstrated poor results on task A, where it failed to correctly identify the title block in 6 images out of 10. By inputting the entire drawing and asking GPT-4o to return the text from the title block, the output included extra text coming from the main content of the drawing, outside of the title block; additionally, the text of the title block itself was incorrect, incomplete or noisy in places (see Figure 4). This is likely due to how the GPT4-o pipeline treats high resolution images by applying tiling and resizing.

Consequently, GPT4-o was considered for task (B) alone. The user prompt was refined to: *What text is in this drawing title block? Identify the title block cells and return their content in a JSON format, where the key is the cell title and the value is the content of the cell without the key.* As image input, manually-cropped title blocks were provided. The text extraction resulted perfect, as shown in Figure 5. The extracted text was now limited to the title block content, without content sourced from outside of it, and the text itself is also perfectly accurate.

These very positive results of GPT-4o on task B were convincing enough to try a combined approach. In fact, it is possible to formalise task A as an object detection task, whose output could be fed to GPT-4o to perform task B. A Faster RCNN model was fine-tuned on an small prototypical dataset and it gave satisfactory results on task A. Consequently, an ad-hoc labelled dataset was devised to test the overall pipeline.

**Dataset definition**

Given the conclusions from the previous section, a main focus of this work went into building a dataset of technical engineering drawings that could be used for task A (title block detection) and that can also be extended to future use cases. The dataset currently includes 1385 engineering drawings and 4 categories: *title block, main content, legend*, and *notes*, as shown in 1. The categories have been defined with the main title block detection use case in mind, but with additional categories useful for future work (*Conclusions, limitations and future work*). In general, the drawings are sourced from different engineering disciplines (e.g. structural, mechanical, electrical engineering, etc.), which requires at least a basic domain expertise for each discipline in order to correctly label it. For this reason, the categories were co-defined with domain experts and then the dataset was labelled by a chosen set of discipline experts. Currently, the majority of drawings in the dataset covers the facade engineering discipline (drawings related to building facades), but it already contains examples for other engineering disciplines such as Mechanical, Electrical, Plumbing (MEP), structural engineering, and also general architectural drawings.

In order to label the drawings, and enable the dataset to be easily extensible, a workflow that could be embedded into the daily practice of engineers was devised. In the AEC industry, one of the most common tools for drawing markup is Bluebeam Revu, used for annotating building drawings. It allows users to mark up drawings with comments and annotations, create layers and shapes, and it is commonly employed in drawing revision and review. We enabled engineers to label the drawings using Revu by creating a converter tool between Revu's PDF annotations and COCO (Common Objects In Context), a commonly used dataset format for CV tasks (cocodataset.org). The COCO dataset is version controlled and kept in sync with the annotated PDF dataset.

This dataset labelling and production workflow allows for easy extensibility, version control, and flexibility. It has an advantage over other CV annotation workflows, such as Azure AIDI (Azure (2025)) as it uses an environment already familiar to engineers and enables the flexibility of producing data usable outside of proprietary platforms.

**Title block detection**

The obtained COCO dataset was then used to fine-tune a Faster R-CNN model. The FasterRCNN model was selected for title block detection primarily due to the lightness and efficiency of the model, although larger model architectures, such as YOLO, are currently being explored. MobileNetV3 was selected as a feature extractor; it consists of a hardware-aware network architecture search (NAS) and NetAdapt Yang et al. (2018) algorithm. Using MobileNetV3 enables a fast-responsive FasterRCNN for our pipeline. The feature maps extracted from MobileNetV3 are processed by the regional proposal network (RPN). The RPN proposes multiple regions where targeted objects might be located. It includes a classifier and a regressor: the classifier determines the probability of a proposed region containing targeted objects, while the regressor obtains the coordinates of these regions. The RPN is robust to translational variants, meaning it can effectively handle changes in the position of objects within an image, making it effective for building drawings that involve title block rotation and shape changes. The model fine-tuning was done using an Stochastic Gradient Descent (SGD) optimiser, with a learning rate (LR) of 0.001 and a momentum of 0.9, over 20 training epochs.

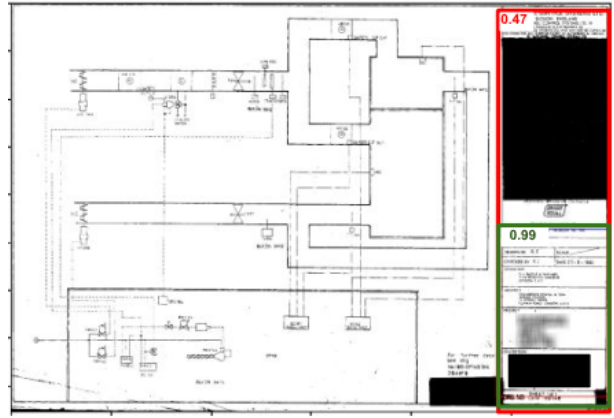

*Figure 6: Examples of TPs and FPs for the title block label and related confidence score. The confidence score is shown in each box top-left corner as a coloured number. Some contents are hidden by black boxes due to data protection.*

The accuracy was determined via Intersection over Union (IoU) of the prediction (P) with the ground truth (GT) bounding box. A prediction P is considered a True Positive (TP) if it predicts the correct label (title block) with an IoU over a threshold of 70%. Redundant True Positives (TPs) for the title block category were ignored, only the topmost-confident prediction is considered. False Positives (FPs) are ignored TP was determined for the same area. This ensures accuracy in reflecting the desired behaviour of the model, based on the assumption that each building drawing contains only one title block. Figure 6 shows examples of TPs and FPs.

Table 1 shows the results of Faster R-CNN detection where the accuracy for title blocks is 0.971. The accuracies for the other labels is lower but still satisfactory. This is likely due to transposition invariance but can be resolved by data augmentation with rotation of non 90-degrees orientation in future work.

*Table 1: Detection results on title block and other categories.*

| Category | Acc | Pr | Rec | F1 |
|---|---|---|---|---|
| **TitleBlock** | **0.971** | **0.84** | **0.96** | **0.83** |
| Main content | 0.943 | 0.82 | 0.954 | 0.84 |
| Notes | 0.82 | 0.40 | 0.93 | 0.32 |
| Legend | 0.86 | 0.54 | 0.95 | 0.41 |

**Title block IE and combined inference pipeline**

To perform IE, an instance of GPT-4o was deployed to receive the detected title block. The system and user prompt used were the same as shown at the end of the *Experiments with existing methods* section.

The combined inference pipeline is illustrated in Figure 7. A drawing is input to the proposed FasterRCNN (A), which detects different components (title blocks, horizon-tal and vertical sections, GA sections, etc.) through a region proposal network (RPN) (B). Detected title blocks are input to GPT-4o, which returns a structured JSON text containing the extracted title block information, as in Figure 5.

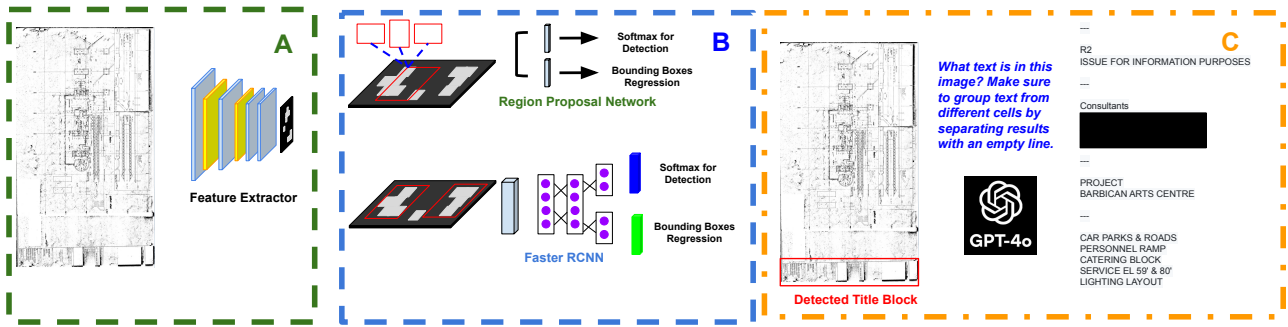

*Figure 7: Overall pipeline.*

The extracted JSON text is then processed to enable result evaluation and better textual search. Importantly, the text extracted from a certain title block cell may not have a specific key (cell title) associated with it (see e.g. the first example in Figure 2); in these cases, GPT-4o *guesses* the cell title, and consistency is not guaranteed. For this reason, semantic text processing for the keys was implemented, using a combination of heuristics (domain-expert-defined synonym dictionary) and small language models (SpaCy, https://spacy.io/). This allowed to reliably detect key-value pairs that represent the same type of metadata across different title blocks and drawings. For example: "drawing description", "description" and "dwg. desc." will all be recognised as the same type of information. Semantic processing was also performed certain cell values, such as the ones containing dates, allowing to uniform textual dates expressed in different formats (i.e. "May 1973", "10/1973") for easier retrieval and comparison.

The evaluation of the extracted JSON text was done by manually extracting text and composing a JSON dictionary from a sample of 50 drawings and comparing it with the text extracted by the pipeline. The evaluation compared the texts using a Levenshtein-based (fuzzy) distance method and was done on the keys and the values. The expected keys were found with an accuracy of 86%, using a fuzzy matching with max distance 2 on keys longer than 10 characters or that include a whitespace. On the matching keys, the values corresponded with an accuracy of 95% using a max fuzzy distance of 4 on values with more than 20 characters.

**User Interface, deployment and efficacy**

The extracted title block information is treated as *metadata* and paired to the drawings to enable search. Initially, the metadata was embedded within the company's file organiser (SharePoint) to enable textual search within an existing system. However, this proved limiting for users, who needed more complex search. Figure 8 shows a prototype of a simple UI that was developed to enable more advanced metadata search. The UI enables conditional search, for example, the user can search `["cinema", "electric"] in "description" AND "date" < 12/1970`, which will give a result where all building drawings contain cinema and electric layouts produced before December 1970. Additionally, the UI offers file renaming and grouping functionality based on the metadata.

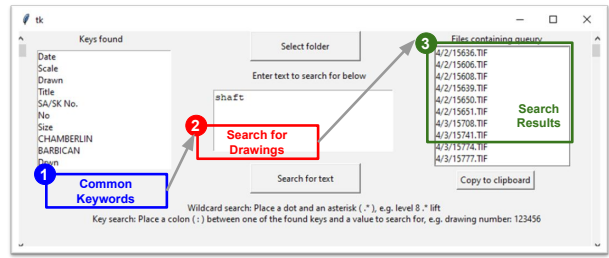

*Figure 8: UI enabling drawing search. 1) shows the keywords found by the analysis, 2) is the search box, and 3) shows the retrieved drawings..*

The pipeline and the UI were successfully deployed on a variety of projects. User analytics proved valuable time and cost-savings, cutting the time spent searching for drawing of at least 40%, reaching 80% on some projects. Importantly, the inference pipeline needs to be run only once per project as the results can stored in a database or paired to the drawings as metadata, so the costs of the GPT-4 instance are negligible when compared to the time savings.

**Conclusions, limitations and future work**

This paper presents a novel title block detection and IE pipeline to enable advanced building drawings search. First, a novel pipeline to allow engineers to efficiently annotate drawings using familiar environment was devised. This pipeline was utilised to produce a labelled dataset. Then, a FasterRCNN model was fine-tuned on this dataset and optimised to detect title blocks in drawings. This model is then included in an inference pipeline that passes the title blocks detected from FasterRCNN to GPT-4o, which performs text extraction in the form of structured

JSON data. Finally, the extracted JSON is treated as drawing metadata and a novel UI enables advanced metadata, allowing drawing retrieval based on text search.

This work overcomes the limitations of low performance on scanned and photographed drawings in previous work such as Li et al. (2024); Kang et al. (2019); Kashevnik et al. (2023), does not require a large dataset for training compared with Shen et al. (2020); Smock et al. (2022); Nassar et al. (2022); Ly and Takasu (2023), and does not rely on CV models that are heavy or expensive to fine-tune, leveraging the lightweight FasterRCNN (Ren et al. (2016)). Additionally, the proposed data annotation pipeline can enable domain experts to efficiently label drawings with virtually no learning curve. Finally, this work proves that utilizing state-of-the-art language models can facilitate IE and enable efficient methods for AEC drawings retrieval.

However, this work focuses on the title block and does not consider the main drawing content for IE and drawing search. Other works such as Van Daele et al. (2021) proved the usefulness of neuro-symbolic IE approaches combining the understanding of both the title block and the main drawing content. Limitations of such works is that they are applied to vector (CAD) drawings, which are often cleaner, simpler and tidier if compared to historical building drawings. Therefore, future work will investigate the design of a domain-specific knowledge graph (KG) to represent the drawing information and connect the labelled categories (Ding and Zhou (2025)). Additionally, work will be carried to define a set of drawing symbols to expand the dataset categories. The selection of symbols will be largely driven by use cases and project demand, with a coordinate effort to embed them in a unified KG. By capturing relationships between the different drawing elements (Figure 1) and drawing symbols, it will be possible to perform neuro-symbolic approaches, and graph deep learning techniques will be explored, leveraging the developed KG. The intention is to unlock novel IE workflows that require an enhanced contextual understanding of building drawings, allowing to cover additional use cases, such as compliance checking, error detection and image-based search, and flexible archiving, ultimately improving the ability to categorize, search, verify and interpret engineering drawings with unprecedented efficiency and efficacy.

## Acknowledgments

This research is supported by the Innovate UK Knowledge Transfer Partnerships (KTP) program (P.no.: 10089005, ref: 13864). Special thanks to Stephan Wassermann-Fry and Masaki Hattori from Buro Happold for their support.